\documentclass[10pt,twocolumn,letterpaper]{article}

\usepackage{cvpr}              

\usepackage{graphicx}
\usepackage{amsmath}
\usepackage{amssymb}
\usepackage{booktabs}
\usepackage{bbm}
\usepackage{multirow}
\usepackage{subcaption}

\makeatletter
\def\hlinew#1{%
	\noalign{\ifnum0=`}\fi\hrule \@height #1 \futurelet
	\reserved@a\@xhline}
\makeatother%
\usepackage[pagebackref,breaklinks,colorlinks]{hyperref}

\usepackage[capitalize]{cleveref}
\crefname{section}{Sec.}{Secs.}
\Crefname{section}{Section}{Sections}
\Crefname{table}{Table}{Tables}
\crefname{table}{Tab.}{Tabs.}

\def\cvprPaperID{1828} 
\def\confName{CVPR}
\def\confYear{2022}

\begin{document}

\title{Unpaired Deep Image Deraining Using Dual Contrastive Learning}

\author {
	Xiang Chen \textsuperscript{\rm 1,2},
	Jinshan Pan \textsuperscript{\rm 2},
	Kui Jiang \textsuperscript{\rm 3},
	Yufeng Li \textsuperscript{\rm 1},
	Yufeng Huang \textsuperscript{\rm 1}, \\
	Caihua Kong \textsuperscript{\rm 1},
	Longgang Dai \textsuperscript{\rm 1},
    Zhentao Fan \textsuperscript{\rm 1} \\
\textsuperscript{\rm 1} Shenyang Aerospace University, \textsuperscript{\rm 2} Nanjing University of Science and Technology, \textsuperscript{\rm 3} Wuhan University \\
\tt \small{https://cxtalk.github.io/projects/DCD-GAN.html}
}
\maketitle

\begin{abstract}
Learning single image deraining (SID) networks from an unpaired set of clean and rainy images is practical and valuable as acquiring paired real-world data is almost infeasible. However, without the paired data as the supervision, learning a SID network is challenging. Moreover, simply using existing unpaired learning methods (\eg, unpaired adversarial learning and cycle-consistency constraints) in the SID task is insufficient to learn the underlying relationship from rainy inputs to clean outputs as there exists significant domain gap between the rainy and clean images. In this paper, we develop an effective unpaired SID adversarial framework which explores mutual properties of the unpaired exemplars by a dual contrastive learning manner in a deep feature space, named as DCD-GAN. The proposed method mainly consists of two cooperative branches: Bidirectional Translation Branch (BTB) and Contrastive Guidance Branch (CGB). Specifically, BTB exploits full advantage of the circulatory architecture of adversarial consistency to generate abundant exemplar pairs and excavates latent feature distributions between two domains by equipping it with bidirectional mapping. Simultaneously, CGB implicitly constrains the embeddings of different exemplars in the deep feature space by encouraging the similar feature distributions closer while pushing the dissimilar further away, in order to better facilitate rain removal and help image restoration. Extensive experiments demonstrate that our method performs favorably against existing unpaired deraining approaches on both synthetic and real-world datasets, and generates comparable results against several fully-supervised or semi-supervised models.
\end{abstract}

\section{Introduction}
Images captured under complicated rain weather environments often suffer from unfavorable visibility by rain streaks. Such these degraded images usually affect many computer vision tasks (including detection \cite{liu2020deep}, segmentation \cite{wojna2019devil} and video surveillance \cite{sultani2018real}, etc.) with drastic performance drop. Thus, it is of great interest to develop an effective algorithm to recover high-quality rain-free images. 

In general, the rain process is usually modeled by the following linear superimposition model:
\begin{equation}
	O = B+R,
	\label{eq: rain-model}
\end{equation}
where $O$, $B$, and $R$ denote the rainy image, clean image, and rain streaks, respectively. The goal of single image deraining (SID) is to estimate clean image $B$ from the input rainy image $O$. This is an ill-posed problem as only rainy image is known. To make the problem well-posed, conventional methods \cite{kang2011automatic, luo2015removing, li2016rain, zhang2017convolutional} usually impose certain priors on the clean images and rain components. 
Although decent results have been achieved, the priors are based on empirical statistical results and may not model the inherent properties of the clean images and rain components, which thus do not effectively remove the rain well.

\begin{figure}[!t]
	\centering
	\includegraphics[width=1.0\columnwidth]{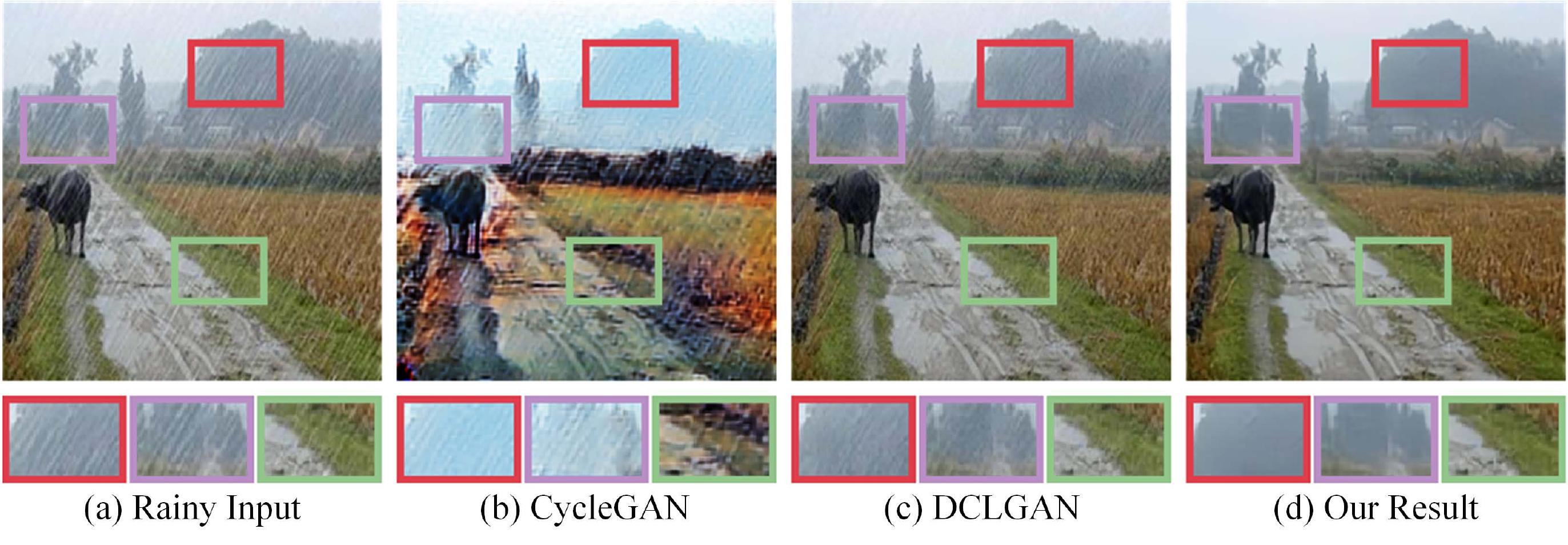}
	\vspace{-6mm}
	\caption{Derained results on a real-world rainy image by unpaired learning approaches. Directly employing existing unpaired learning methods, \eg, CycleGAN and DCLGAN, does not effecively remove rain streaks.}
	\label{fig1}
	\vspace{-3mm}
\end{figure}

Recently, numerous data-driven learning methods have developed \cite{ren2019progressive, zhang2019image, jiang2020multi, yang2020towards, yi2021structure}.
Although remarkable performance has been achieved, these fully-supervised methods need paired synthetic data which does not model the real-world degradation well. Therefore, these methods usually do not perform well when handling the real-world rainy images due to the domain gap between the training and test data.
Moreover, obtaining large scale paired real-world data for training in complex rainy environments is challenging. 

To overcome these problems, the semi-supervised \cite{wei2019semi, yasarla2020syn2real, huang2021memory,liu2021unpaired} and unsupervised \cite{zhu2019singe, han2020decomposed} learning have been proposed for SID.
These methods \cite{jin2019unsupervised, huang2021memory} either focus on domain invariant features by leveraging the limited labeled data and introducing the auxiliary optimization objectives or develop domain adaption strategies \cite{wei2021deraincyclegan, han2020decomposed} to improve the generalization capabilities of the deep models. 
However, without suitable constraints for rain streaks and clean images, existing unpaired learning methods \cite{zhu2017unpaired, han2021dual} do not effectively restore high-quality derained results, as shown in Fig. \ref{fig1} (b) and (c).
Note that, these methods mainly consider the mapping relationship in the image space but ignore the potential relationship in the feature space, which does not fully excavate the useful feature information for SID.
Since the ground truth labeled data is not fully available, how to model the latent-space representation by exploring the relationship between the rainy inputs and clean outputs is important for the deep learning-based methods. 
In addition, given that clean images can be easily obtained, it is also important to develop an effective method that can explore properties of the clean exemplars to facilitate image restoration when paired data is not available. 

Towards this end, we develop a Dual Contrastive Derain-GAN (DCD-GAN) method which incorporates recent contrastive learning with adversarial framework to explore useful features from the rainy images and unpaired clean images so that the extracted features can better facilitate rain removal.
Ideally, we note that if a deep model can accurately restore a clean image from rainy one, the features that are for clean image reconstruction would have mutual information with the ones from the ground truth rain-free images by the same deep model.
This motivates us to introduce a contrastive learning method to mining the mutual features of rainy images and clean ones in the deep feature space, so that we can use the features from the clean images to better guide the image restoration. 
Therefore, our proposed DCD-GAN includes two main interactive branches: Bidirectional Translation Branch (BTB) and Contrastive Guidance Branch (CGB). 
Specifically, BTB is equipped with bidirectional mapping to mine rain-related or clean-cue features, and the use of adversarial training produces rich exemplar pairs during the optimization. 
In addition, CGB implicitly constrains the latent space of corresponding patches to guide deraining by encouraging the positives (i.e, similar feature distributions) closer while keeping the negatives (i.e, dissimilar ones) further away. 
To summarize, the main contributions of this paper are summarized as follows:

\begin{itemize}
	\item We formulate an effective DCD-GAN which leverages dual contrastive learning to encourage the model to explore mutual features while distinguish the dissimilar ones between the rainy domain and the rain-free domain in the deep feature space.
	\item The proposed DCD-GAN is performed without paired training information, where the features from the unpaired clean exemplars can facilitate rain removal. The learned latent restoration can boost the cross-domain deraining generalization performance.
	\item Experimental results on challenging datasets considerably show that our developed method performs favorably against existing unpaired deraining approaches, and achieves comparable performance against several fully-supervised or semi-supervised models.
\end{itemize}	

\section{Related Work}

\subsection{Single Image Deraining}
Existing deep learning-based SID methods are categorized into paired (fully-supervised), semi-supervised and unpaired (without paired supervised) approaches \cite{yang2020single}.

\textbf{For the paired deraining methods}, Fu {\em et al.} \cite{fu2017clearing} first employ the Derain Net with multi-layer CNN to extract and remove the rain layer, and further introduce deep detail network (DDN) \cite{fu2017removing} that directly removes the rain streaks by reducing the mapping range. Later, Zhang {\em et al.} \cite{zhang2018density} present a density-aware multi-stream densely connected CNN algorithm, termed DID-MDN, to characterize rain streaks by estimating rain density. The classic work of RESCAN \cite{li2018recurrent} introduces a convolutional and recurrent neural network-based way to make full use of the contextual information. Spatial attentive network (SPANet) \cite{wang2019spatial} captures the spatial contextual information based on the recurrent network and applies a branch to obtain the spatial details in a local-to-global manner. By replacing low-quality features by latent high-quality features, a robust representation learning network is proposed by \cite{chen2021robust} to address deraining model errors.

\begin{figure*}[t]
	\centering
	\includegraphics[width=0.90\textwidth]{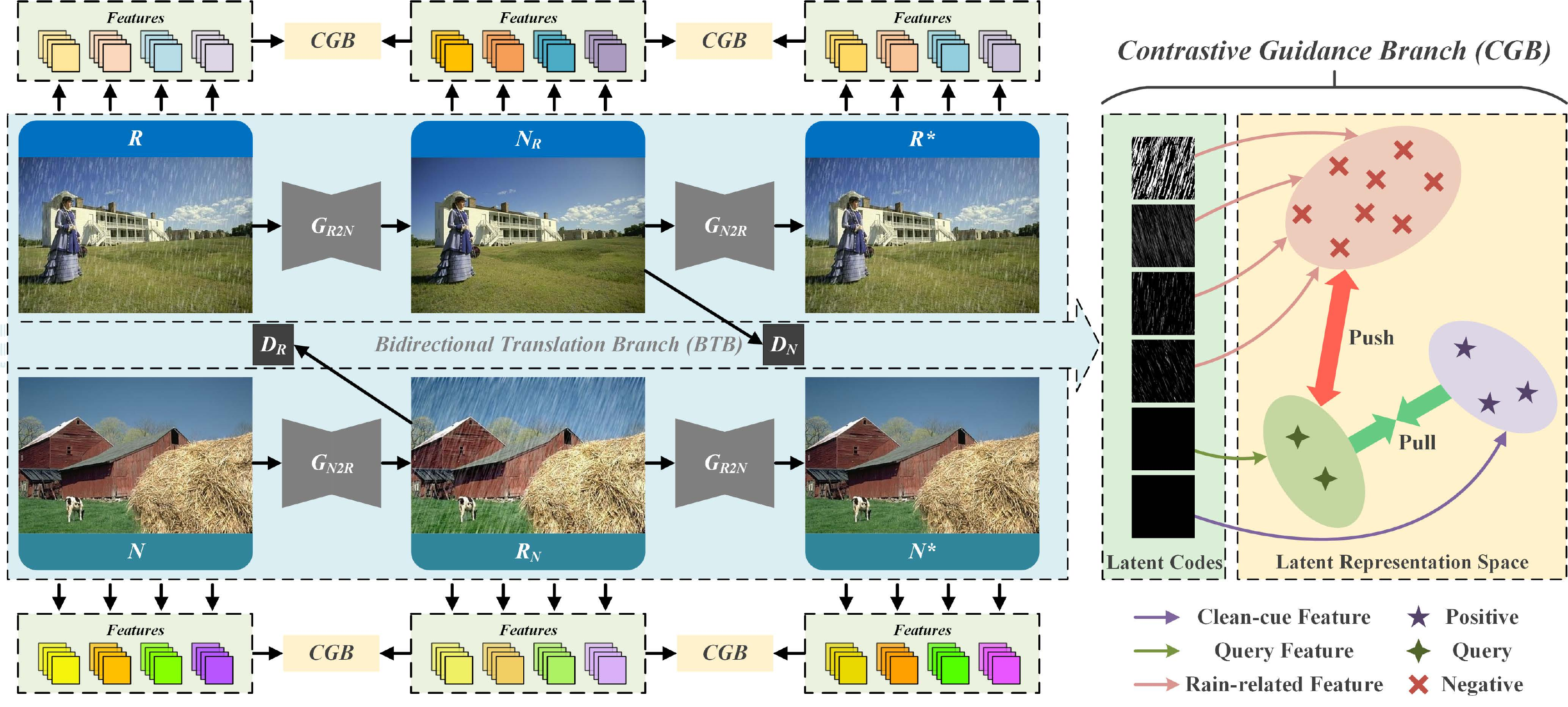}
	\vspace{-2mm}
	\caption{The overall framework of our Dual Contrastive Derain-GAN (DCD-GAN). There are two cooperative branches, bidirectional translation branch (BTB) and contrastive guidance branch (CGB). In the feature space, images generated from close (“positive”) latent codes are visually similar, while images generated from far-away (“negative”) latent codes are visually disimilar. Our developed CGBs aim to learn a representation to pull similar feature distribution (\eg, clean $\rightarrow \leftarrow$ clean) and push disimilar (\eg, rain $\leftarrow \rightarrow$ clean) apart.}
	\label{fig2}
    \vspace{-3mm}
\end{figure*}

\textbf{For the semi-supervised learning-based methods}, Wei {\em et al.} \cite{wei2019semi} utilize a semi-supervised learning framework (SSIR) to analyze the residual difference between domains. The work of Syn2Real \cite{yasarla2020syn2real} develops a Gaussian process-based semi-supervised method, which can increase the network ability by using synthetic images and unlabeled real rainy images. In \cite{liu2021unpaired}, a rain direction regularizer is designed to constrain the network by combining the semi-supervised learning part and the knowledge distillation part.

\textbf{For the unpaired deraining methods}, CycleGAN \cite{zhu2017unpaired} has been widely employed to address the unpaired image-to-image translation issue, to achieve deraining process. 
Recent works \cite{zhu2019singe,wei2021deraincyclegan,jin2019unsupervised,han2020decomposed} all use the improved CycleGAN architecture and constrained transfer learning to jointly the rainy and rain-free image domains. As mentioned before, it is difficult for these methods to learn accurate transformation between two domains using only cycle-consistency constraints, since their domain knowledge is asymmetrical. Instead, our approach can overcome this limitation and incorporates contrastive learning with adversarial training to further improve deraining robustness in unpaired setting.

\subsection{Contrastive Learning}
Contrastive learning has witnessed a significant progress in the field of self-supervised and unsupervised representation learning. Instead of using fixed targets and pre-defined ones, contrastive regularization maximizes the mutual features between different domains by exploiting both the information of positive pairs and negative pairs. Specifically, it aims to learn suitable embeddings by pulling the exemplar close to positive samples while pushing it far away from negative samples in the related representation space. Recent studies have applied contrastive losses in a host of low-level vision tasks and achieved SOTA performance, \eg, image dehazing \cite{wu2021contrastive}, image denoising \cite{dong2021residual}, image super-resolution \cite{zhang2021blind} and underwater image restoration \cite{han2021single}. Similarly, the recent work of DCLGAN \cite{han2021dual} takes the advantages of both CUT \cite{park2020contrastive} and CycleGAN \cite{zhu2017unpaired} to cope with unsupervised image-to-image translation problem. Different from its mutual mapping in complex image space, we show the use of contrastive constraints to perform unpaired learning in deep feature space, which can better help image restoration.

\section{Proposed Method}
Fig. \ref{fig2} presents the overall architecture of our proposed DCD-GAN. It consists of BTB and CGB branches. BTB is used to guide transfer ability and latent feature distributions between the rainy domain and the rain-free domain. Furthermore, CGBs are adopted to constrain the mutual feature information between two domains, as well as the multiple loss functions to provides jointly optimization. We describe the details about the proposed network in the following. 

\subsection{Bidirectional Translation Branch (BTB)}
Let $\mathbb{D}_{R}$ denotes the set of the rainy images without ground truth labels and $\mathbb{D}_{N}$ denotes the set of clean exemplar images, our final goal is to learn a deep model to explore the features from these unpaired data $\mathbb{D}_{R}$ and $\mathbb{D}_{N}$ without the supervision of the ground truth labels to estimate the rain-free images. To this end, we introduce a bidirectional translation branch (BTB) as the backbone of the proposed DCD-GAN. Specifically, BTB contains two generators $G_{R 2 N}$, $G_{N 2 R}$ generating the clean and rainy images respectively and two discriminators $D_{R}$, $D_{N}$ distinguishing between fake derained images and real clean images. Continuing along the trajectory of \cite{wei2021deraincyclegan, ye2021closing}, the overall pipeline of BTB covers two circuits of rain generation and rain removal: (1) rainy to rainy forward cycle-consistency transformation $R \rightarrow N_{R} \rightarrow R^{*}$ ; (2) rain-free to rain-free backward cycle-consistency transformation $N \rightarrow R_{N} \rightarrow N^{*}$.

Due to the circulatory architecture of bidirectional learning, the exemplars generated by BTB could be mined rain-related or clean-cue features during the optimization, thus forming latent feature distributions to explore regularization. Notably, unpaired clean exemplars from the backward cycle can facilitate rain removal. We will show the influence of the clean exemplars in Section \ref{4.3}. In fact, it will be not sufficient to utilize cycle-consistency to remove rain, in that this constraint is weak. Hence, we develop contrastive constraints to formulate multiple contrastive guidance branches (CGBs) that serve as internal bridges into the BTB, yielding more faithful solutions (good generation and good reconstruction). The CGB component is described below.

\begin{figure}[!t]
	\centering 	
	\begin{subfigure}[t]{0.47\columnwidth}
		\centering
		\includegraphics[width=\columnwidth]{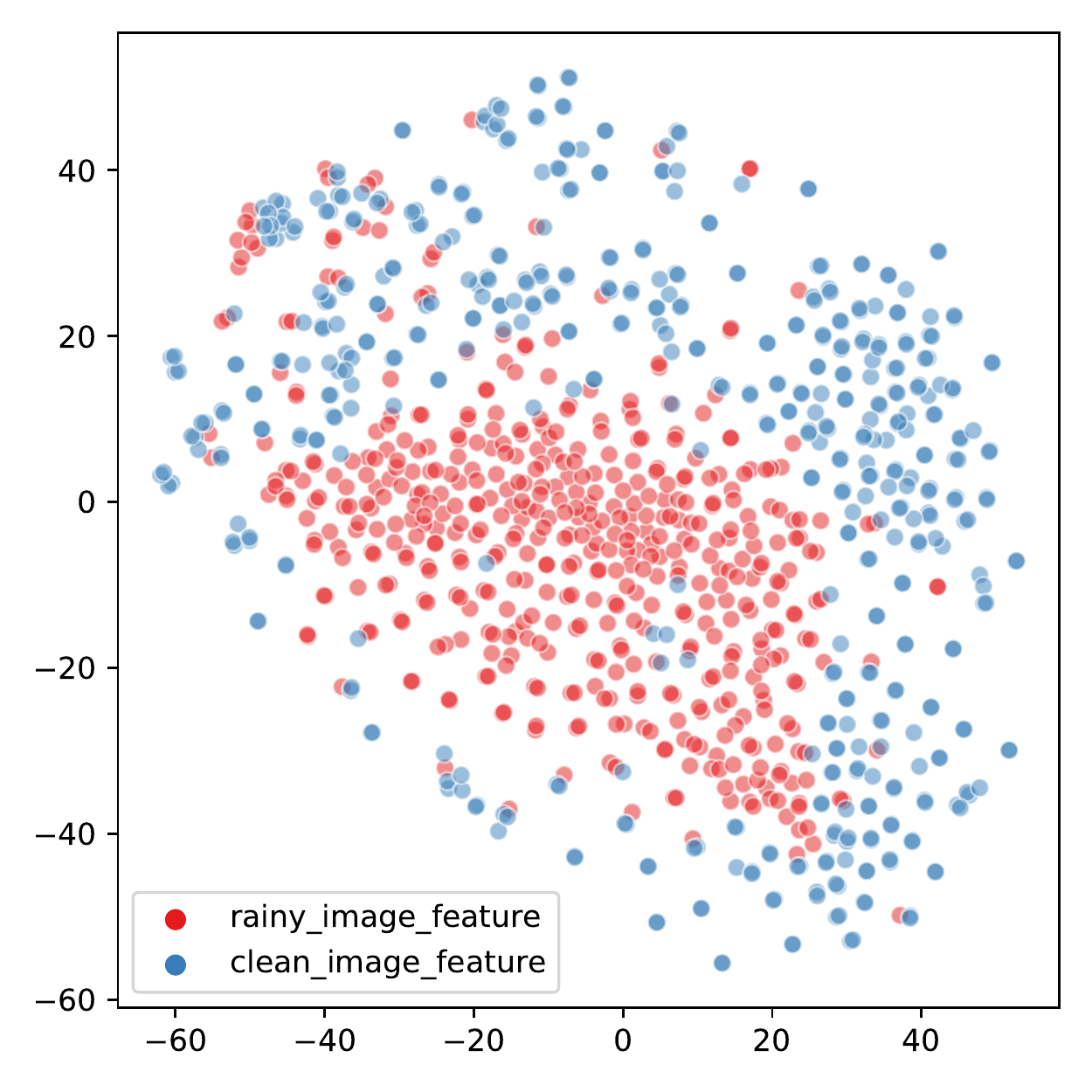}
		\caption{w/o CGB}
	\end{subfigure}
	\begin{subfigure}[t]{0.47\columnwidth}
		\centering
		\includegraphics[width=\columnwidth]{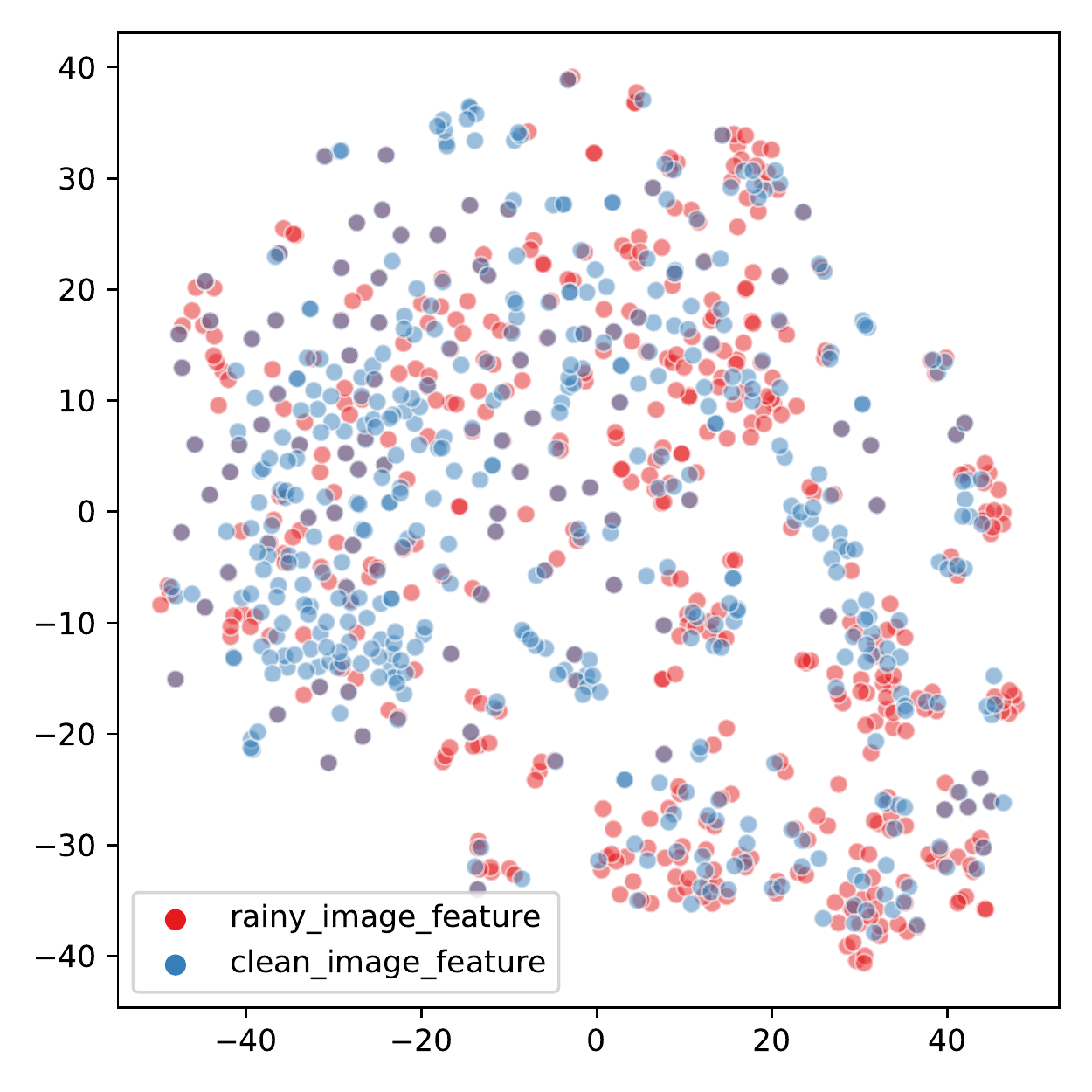}
		\caption{w CGB}
	\end{subfigure}
	\vspace{-3mm}
	\caption{The t-SNE visualization of features learned with/without our CGB. The red round point denotes the deep features extracted from rainy images for clean image reconstruction. The blue round points denote the deep features extracted from clean images. With CGBs, the embeddings of the features from the clean image and the rainy image have low distances in the latent space. Thus, using the rainy features by CGBs is able to generate clean images.} 
	\label{fig3} 
	\vspace{-3mm}
\end{figure} 

\subsection{Contrastive Guidance Branch (CGB)}
To guide in better deraining, CGBs are involved between $\mathbb{D}_{R}$ and $\mathbb{D}_{N}$ to exploit advantage of the learned domain-relevant feature distribution. Our primary insight is that the features extracted from $\mathbb{D}_{R}$ share some mutual properties to the features extracted from $\mathbb{D}_{N}$. If pairs of patches generated from different domains with low feature disparity are similar (referred to as “positive”), their embeddings in latent space should also have a low distance, and vice-versa. Therefore, we propose to learn the image restoration in the deep feature space by enforcing contrastive constraints.

Following the advantage of dual learning setting in stability training \cite{yi2017dualgan}, we take the auxiliary encoder $E$ as separate embeddings to extract features from domain $\mathbb{D}_{R}$ and $\mathbb{D}_{N}$, respectively. As suggested in \cite{liu2021divco,park2020contrastive}, by using off-the-shelf generators to extract feature representations in contrastive learning, excessive modifications to the original architecture can be avoided. Consequently, we directly extract features of images from the $L$ encoding layers of the generator ($G_{R 2 N}$ and $G_{N 2 R}$) as the auxiliary encoder of CGB and send them to a two-layer multi-layer perceptron (MLP) projection head.  Here, we do not share weights, thus capturing variability in both domains and learning better embeddings. After having different stacks of features, our CGB constrains the contrastive distribution learning with two distinct embeddings, with the goal of associating a query generated exemplar and its similar feature distribution (\eg, clean $\rightarrow \leftarrow$ clean) of the representations, in contrast to other dissimilar ones (\eg, rain $\leftarrow \rightarrow$ clean) at the same time for implicitly supervising the deraining process.

To constrain mutual features in their embeddings, let us define the query code $\hat{x}$ sampled from one domain, the positive code $\hat{x}^{+}$ and $k$ negative codes $\left\{\hat{x}_{i}^{-}\right\}_{i-1}^{k}$ sampled from the other domain. Different from previous methods \cite{park2020contrastive,wu2021contrastive} that only select negative samples within the input image, we focus on both internal and external latent codes concatenating the two sides of the CGB within the feature space, and experiments demonstrate our strategy achieves better benefits for SID (see Section \ref{4.3}). The constraint of CGB is enforced by encouraging the positives closer while keeping the negatives further away, so as to help the deraining generators guide correctly in transforming the rainy inputs to the clean outputs. To understand the effect of such CGBs, we further use t-SNE \cite{van2008visualizing} to visualize learned features in Fig. \ref{fig3}. Ideally, if a deep model can accurately restore a clean image from rainy one, the features that are for clean image reconstruction and the feature of ground truth rain-free images will completely overlap in the deep feature space. Obviously, with CGBs, these two types of features are pulled together, which become well distinguished and able to capture the relationship between $\mathbbm{D}_{R}$ and $\mathbbm{D}_{N}$. Mathematically, these corresponding feature representations are then formalized as: $f=E(\hat{x})$, $f^{+}=E\left(\hat{x}^{+}\right)$, $f_{i}^{-}=E\left(\hat{x}_{i}^{-}\right)$. Thus, the \textbf{Contrastive Loss} is developed to regularize the captured images’ feature distributions, which can be written as:
\begin{equation}
	\begin{aligned}
		&L_{c o n t}\left(G_{R 2 N}, G_{N 2 R}\right)= \\
		&\mathbbm{E}_{r \sim R, n \sim N}\left[-\log \frac{\operatorname{sim}\left(f, f^{+}\right)}{\operatorname{sim}\left(f, f^{+}\right)+\sum_{i=1}^{N} \operatorname{sim}\left(f, f_{i}^{-}\right)}\right],
	\end{aligned}
\end{equation}
where $\operatorname{sim}(u, v)=\exp \left(\frac{u^{T} v}{\|u\|\|v\| \tau}\right)$ is the cosine similarity function that computes the similarity between two normalized feature vectors, and $\tau$ denotes a temperature parameter. 

\begin{table*}[t]\footnotesize
	\centering
	\caption{Comparison of quantitative results on four benchmark datasets. Bold and \textit{italic} indicate the best and second-best results.}
	\vspace{-3mm}
	\begin{tabular}{c|c|cc|cc|cccc}
		\hlinew{1.0pt}
		\multicolumn{2}{c|}{Datasets}                                                                                           & \multicolumn{2}{c|}{\textit{Test100}} & \multicolumn{2}{c|}{\textit{Test1200}} & \multicolumn{2}{c|}{\textit{Test1400}}               & \multicolumn{2}{c}{\textit{Test1000}} \\ \hline
		\multicolumn{2}{c|}{Metrics}                                                                                            & PSNR              & SSIM              & PSNR               & SSIM              & PSNR           & \multicolumn{1}{c|}{SSIM}           & PSNR              & SSIM              \\ \hline
		\multirow{2}{*}{Prior-based methods}                                                                    & DSC \cite{luo2015removing}          & 18.56             & 0.599             & 24.24              & 0.827             & 27.31          & \multicolumn{1}{c|}{0.837}          & 34.95             & 0.941             \\
		& GMM \cite{li2016rain}           & 20.46             & 0.730             & 25.66              & 0.817             & 26.87          & \multicolumn{1}{c|}{0.808}          & 34.30             & 0.942             \\ \hline
		\multirow{4}{*}{\begin{tabular}[c]{@{}c@{}}Paired / Supervised\\ methods\end{tabular}}                  & DDN \cite{fu2017removing}           & 21.16             & 0.732             & 27.93              & 0.853             & 28.00          & \multicolumn{1}{c|}{0.873}          & 34.70             & 0.934             \\
		& DID-MDN \cite{zhang2018density}       & 21.89             & 0.795             & 29.66              & 0.899             & 26.38          & \multicolumn{1}{c|}{0.835}          & 34.68             & 0.930             \\
		& RESCAN \cite{li2018recurrent}        & 24.09             & 0.841    & \textit{32.25}     & 0.907             & \textbf{32.03} & \multicolumn{1}{c|}{\textit{0.917}} & 34.71             & 0.937             \\
		& SPANet \cite{wang2019spatial}       & \textit{24.37}    & \textbf{0.861}    & 30.05              & \textit{0.934}    & 29.83          & \multicolumn{1}{c|}{0.904}          & 35.13   & \textit{0.944}    \\ \hline
		\multirow{2}{*}{Semi-supervised methods}                                                                & SIRR \cite{wei2019semi}         & /                 & /                 & 30.57              & 0.910             & 30.01          & \multicolumn{1}{c|}{0.907}          & 34.85             & 0.936             \\
		& Syn2Real \cite{yasarla2020syn2real}     & 23.74             & 0.799             & /                  & /                 & /              & \multicolumn{1}{c|}{/}              & /                 & /                 \\ \hline
		\multirow{4}{*}{\begin{tabular}[c]{@{}c@{}}Unpaired / Without paired\\ supervised methods\end{tabular}} & CycleGAN \cite{zhu2017unpaired}     & 21.92             & 0.683             & 24.17              & 0.729             & 24.34          & \multicolumn{1}{c|}{0.776}          & 31.59             & 0.896             \\
		& RR-GAN \cite{zhu2019singe}       & 23.51             & 0.757             & /                  & /                 & /              & \multicolumn{1}{c|}{/}              & /                 & /                 \\
    	& DerainCycleGAN \cite{wei2021deraincyclegan} & 24.30             & \textit{0.854}             & /                  & /                 & /              & \multicolumn{1}{c|}{/}              & \textit{35.20}              & \textbf{0.950} \\
	
		& \textbf{Ours} & \textbf{25.61}    & 0.813             & \textbf{32.46}     & \textbf{0.937}    & \textit{31.65} & \multicolumn{1}{c|}{\textbf{0.918}} & \textbf{35.30}    & 0.943    \\ \hlinew{1.0pt}
	\end{tabular}
    \vspace{-3mm}
	\label{table1}	
\end{table*}

\begin{table}[!t]
	\centering
	\caption{Descriptions of different benchmark training datasets.}
	\vspace{-3mm}
	\resizebox{.95\columnwidth}{!}{
		\begin{tabular}{c|ccc|c}
			\hlinew{1.0pt}
			Datasets  & Rain800  & DID-Data & DDN-Data & SPA-Data   \\ \hline
			Train-Set & 700      & 12000    & 12600    & 28500      \\
			Test-Set  & 100      & 1200     & 1400     & 1000       \\ \hline
			Type      & \multicolumn{3}{c|}{Synthetic} & Real-world \\ \hline
			Name      & \textit{Test100}  & \textit{Test1200} & \textit{Test1400} & \textit{Test1000}   \\ \hlinew{1.0pt}
		\end{tabular}
	}
	
	\vspace{-3mm}
	\label{table2}
\end{table}

\subsection{Loss Function}
As the ground truths are not available, it is vital to develop effective loss function to constrain the DCD-GAN for better SID. In addition to the contrastive loss, we further introduce other constraints to regularize the network training.

\textbf{Color Cycle-Consistency Loss}. It is worth noting that previous CycleGAN-based methods mostly adopt the cycle-consistency loss to preserve image content. However, there exists the problem of “channel pollution” \cite{tang2018gesturegan,tang2019asymmetric} in cycle-consistency loss, because generating a whole image at one time makes different channels interact with each other, resulting in unpleasant artifacts in final derained results. To alleviate it, we develop a simple yet effective solution by constructing the consistence loss for each channel independently. Here, we calculate the pixel loss of the red, green and blue channels between the input and the reconstructed image separately, and then sum the three distance losses as the color cycle-consistency loss, which can be expressed as:
\begin{equation}
	\begin{split}
		L_{c o l o r c y c}=\sum_{i \in\{r, g, b\}} \mathbbm{E}_{n \sim N}^{i}\left[\left\|G_{R 2 N}\left(G_{N 2 R}(n)\right)-n\right\|_{1}\right] \\
		+\sum_{i \in\{r, g, b\}} \mathbbm{E}_{r \sim R}^{i}\left[\left\|G_{N 2 R}\left(G_{R 2 N}(r)\right)-r\right\|_{1}\right],
	\end{split}
\end{equation}
where $i$ denotes different channels. $r \in \mathbbm{D}_{R}$, $n \in \mathbbm{D}_{N}$ are a rainy image and an unpaired clean image.

\textbf{Adversarial Loss}. We observed that employing the adversarial loss is able to generate the realistic images. Similar to \cite{zhu2019singe}, the adversarial loss in $\mathbbm{D}_{N}$ for SID is defined as:
\begin{equation}
\begin{split}
L_{a d v}\left(G_{R 2 N}, D_{N}\right)=\mathbbm{E}_{n \sim N}\left[\log D_{N}(n)\right] \\ +\mathbbm{E}_{r \sim R}\left[\log \left(1-D_{N}\left(G_{R 2 N}(r)\right)\right)\right],
\end{split}
\end{equation}
where $G_{R2N}$ minimizes the objective function to make the generated clean images look similar to real samples. In contrast, $D_{N}$ maximizes the loss to distinguish generated clean images and real clean images. Here, we omit same adversarial loss in $\mathbbm{D}_{R}$. The overall adversarial loss is calculated by $L_{a d v}=L_{a d v}\left(G_{R 2 N}, D_{N}\right)+L_{a d v}\left(G_{N 2 R}, D_{R}\right)$.

\textbf{Frequency Loss}. Inspired by \cite{huang2021selective,cheng2021iicnet}, we take advantage of frequency domain and exploit a frequency loss to further preserve the content and structure information of the image:
\begin{equation}
\begin{split}
L_{freq}=\mathbbm{E}_{n \sim N}[\left\|FT\left(G_{R 2 N}\left(G_{N 2 R}(n)\right)\right)-FT(n)\right\|_{2}^{2}] \\
+ \mathbbm{E}_{r \sim R}[\left\|FT\left(G_{N 2 R}\left(G_{R 2 N}(r)\right)\right)-FT(r)\right\|_{2}^{2}],
\end{split}
\end{equation}
where $FT$ denotes one-sided Fourier transform to obtain their corresponding frequency domain.

Based on above considerations, the final loss function is:
\begin{equation}
L_{t o t a l}=\lambda_{1} L_{c o n t}+\lambda_{2} L_{c o l o r c y c}+\lambda_{3} L_{a d v}+\lambda_{4} L_{f r e q},
\end{equation}
where $\lambda_{i}$ is the trade-off weight. In our experiments, we empirically set $\lambda_{1}$ = 2, $\lambda_{2}$ = 1, $\lambda_{3}$ = 1, and $\lambda_{4}$ = 0.1.

\section{Experiments}
\subsection{Experimental Settings}
{\flushleft\textbf{Datasets}.} We use four challenging benchmark datasets including Rain800 \cite{zhang2019image}, DID-Data \cite{zhang2018density}, DDN-Data \cite{fu2017removing}, and SPA-Data \cite{wang2019spatial} with various rain streaks of different sizes, shapes and directions. The detailed descriptions of the used training and testing datasets are tabulated in Tab. \ref{table2}. 

{\flushleft \textbf{Implementation Details}.} The overall architecture of DCD-GAN is based on CycleGAN, a Resnet-based \cite{he2016deep} generator with 9 residual blocks and a PatchGAN \cite{isola2017image} discriminator. During training, we use the Adam optimizer with $\beta_{1}$ = 0.5 and $\beta_{2}$ = 0.999 and the models are trained for total 600 epochs. The model is trained from the scratch for 300 epochs with the learning rate of 0.0001, followed by another 300 epochs with the learning rate linearly decayed to 0. In the CGB, we select 255 internal and 256 external negative samples for one query. The temperature parameter $\tau$ is set to 0.07. We use a batch size of 1 and instance normalization \cite{ulyanov2016instance}. The entire network is performed using PyTorch framework on two NVIDIA Tesla V100 GPUs. It should be noted that all training images are randomly cropped to $256 \times 256$ patches in an unpaired learning manner. To better facilitate contrastive learning, we have implemented specific data enhancement strategies for the above benchmarks.

{\flushleft \textbf{Comparison Methods}.} We compare our method with those of two prior-based algorithms (i.e., DSC \cite{luo2015removing} and GMM \cite{li2016rain}), four paired supervised methods (i.e., DDN \cite{fu2017removing}, DID-MDN \cite{zhang2018density}, RESCAN \cite{li2018recurrent}, and SPANet \cite{wang2019spatial}), two semi-supervised approaches (i.e., SSIR \cite{wei2019semi} and Syn2Real \cite{yasarla2020syn2real}), and three unsupervised deep nets (i.e., CycleGAN \cite{zhu2017unpaired}, RR-GAN \cite{zhu2019singe}, and DerainCycleGAN \cite{wei2021deraincyclegan}). As the code of the methods \cite{zhu2019singe,wei2021deraincyclegan} is not available, we refer to some comparison results presented in their paper. For other methods, if there are no pre-trained models, we retrain them with the implementations provided by authors, otherwise we evaluate them directly with the online available codes. 

{\flushleft\textbf{Evaluation Metrics}.} For the images with ground truth, we evaluate each method by two well-known used metrics: PSNR and SSIM \cite{wang2004image}. For the cases without ground truth, we use no-reference quality metrics NIQE \cite{mittal2012no} and BRISQUE \cite{mittal2012making} to perform quantitative comparison.

\begin{figure*}[t]
	\centering
	\includegraphics[width=1.0\textwidth]{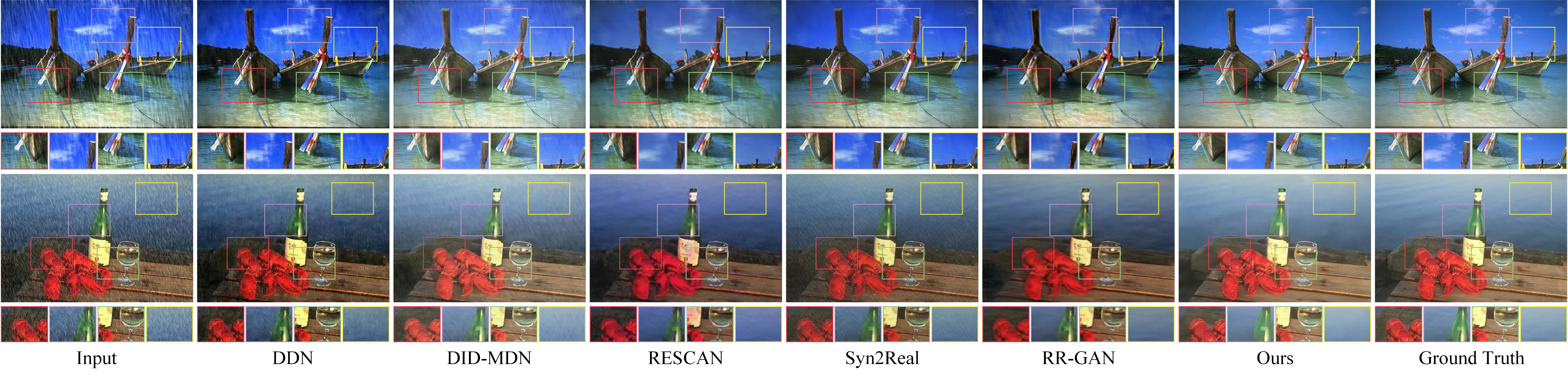} 
	\vspace{-6mm}
   \caption{Qualitative evaluations on the Test100 dataset. Zooming in the figures offers a better view at the deraining capability.}
   	\label{fig4}
	\vspace{-3mm}
\end{figure*}
\begin{figure*}[t]
	\centering
	\includegraphics[width=1.0\textwidth]{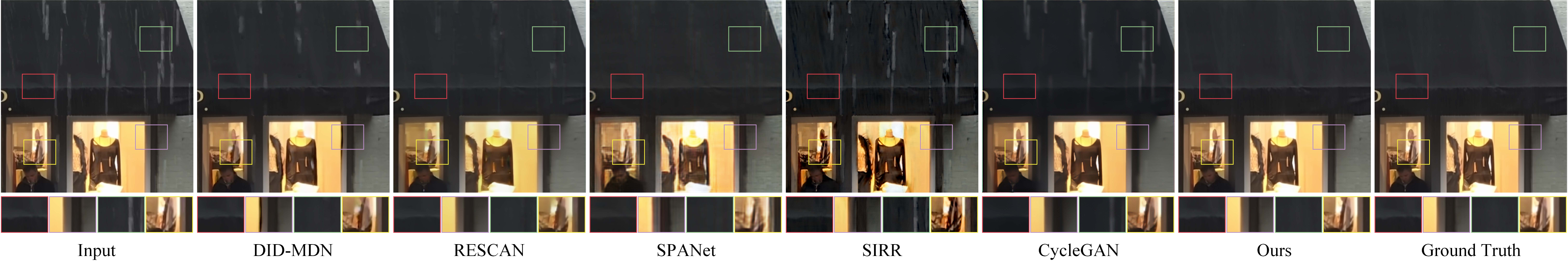} 
	\vspace{-6mm}
	\caption{Qualitative evaluations on the Test1000 dataset. Zooming in the figures offers a better view at the deraining capability.}
	\label{fig5}
	\vspace{-3mm}
\end{figure*}
\begin{figure*}[!t]
	\centering
	\includegraphics[width=0.95\textwidth]{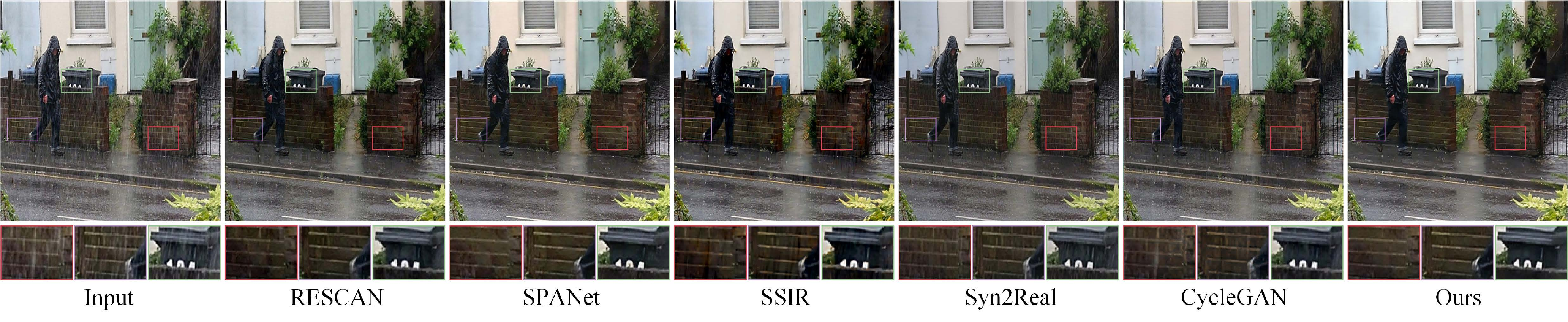} 
	\vspace{-3mm}
	\caption{Qualitative evaluations on a real rainy image. Zooming in the figures offers a better view at the deraining capability.}
	\label{fig6}
	\vspace{-3mm}
\end{figure*}

\subsection{Experimental Results}
{\flushleft\textbf{Synthetic Data}.} Tab. \ref{table1} reports the quantitative results of different methods on synthetic datasets including Test100, Test1200, and Test1400. We can observe that: (1) compared with unpaired deraining methods, our net achieves state-of-the-art results of PSNR, which reflects the excellent performance of DCD-GAN. (2) For semi-supervised methods, there is still a performance gap with fully-supervised models, even worse than our unpaired net. (3) Our approach can deliver competitive results to several existing paired supervised models, thanks to the additional constraints provided by CGBs. Besides, we show some hard examples for visual observation comparisons in Fig. \ref{fig4}. According to the ground truth, all the other works do an undesired work in details and color restoration. In contrast, our developed method not only be able to remove various types of rain streaks but also preserve color and structural properties of underlying objects to a substantial degree.

\begin{table}[t]
	\centering
	\caption{Comparison of quantitative results on real-world rainy images, and note that lower scores indicate better image quality.}
	\vspace{-3mm}
	\resizebox{.95\columnwidth}{!}{
		\begin{tabular}{c|cccccc}
			\hlinew{1.0pt}
			Method  & RESCAN & SPANet & SSIR   & Syn2Real & CycleGAN & Ours           \\ \hline
			NIQE $\downarrow$   & 4.571  & 4.317  & 4.263  & 4.354    & 4.689    & \textbf{4.032} \\
			BRISQUE $\downarrow$ & 25.446 & 20.220 & 27.418 & 15.920    & 23.801   & \textbf{13.311} \\ 	\hlinew{1.0pt}
	\end{tabular} }
	\vspace{-3mm}
	\label{table3}
\end{table}

{\flushleft\textbf{Real-world Data}.} For further general verification in practical use, we conduct comparisons against other algorithms on the real-world rainy dataset. Since SPA-Data \cite{wang2019spatial} has its ground truth by human labeling, it can be evaluated using the numerical metrics. According to the numerical value in the last column of Tab. \ref{table1}, our proposed model of DCD-GAN gets the best values on PSNR metric and obtains competitive SSIM values compared with other related nets on Test1000. From Fig. \ref{fig5}, it can be observed that the proposed method recovers clearer images with faithful structures.

Furthermore, we choose 30 real-world rainy images from Google image search. Fig. \ref{fig6} illustrates the deraining performance of all competing approaches on one sample image. As rainy images from natural scenarios are more complex and challenging, our method still exhibits remarkable performance with less rain streaks residue. In contrast, other approaches fail to achieve the desired results. According the average values of NIQE and BRISQUE in Tab. \ref{table3}, our method gets the lower scores, which means a high-quality output with higher naturalness and better fidelity against other comparative models on real-world images.

\subsection{Ablation Study}
\label{4.3}
{\flushleft\textbf{Effectiveness of Proposed Loss Function}.} In order to evaluate the effectiveness of our hybrid loss function, we conduct an ablation study on Test100. To be specific, we regularly remove one component to each configuration at one time. To ensure the fair comparison, the same training settings are kept for all models testing, except the modification depicted in Tab. \ref{table4}. The best performance achieves 25.61 dB by using all the above component, so that each loss terms we consider has its own contribution in deraining process. We further investigate our color cycle-consistency loss effect by comparing with commonly used cycle-consistency loss ($L_{c y c}$). Through the last column of Tab. \ref{table4}, we can observe that the performance drops with the replacement of color cycle-consistency loss, which confirms its advantages for improving the quality of recovery results.

\begin{figure}[!t]
	\centering
	\includegraphics[width=1.0\columnwidth]{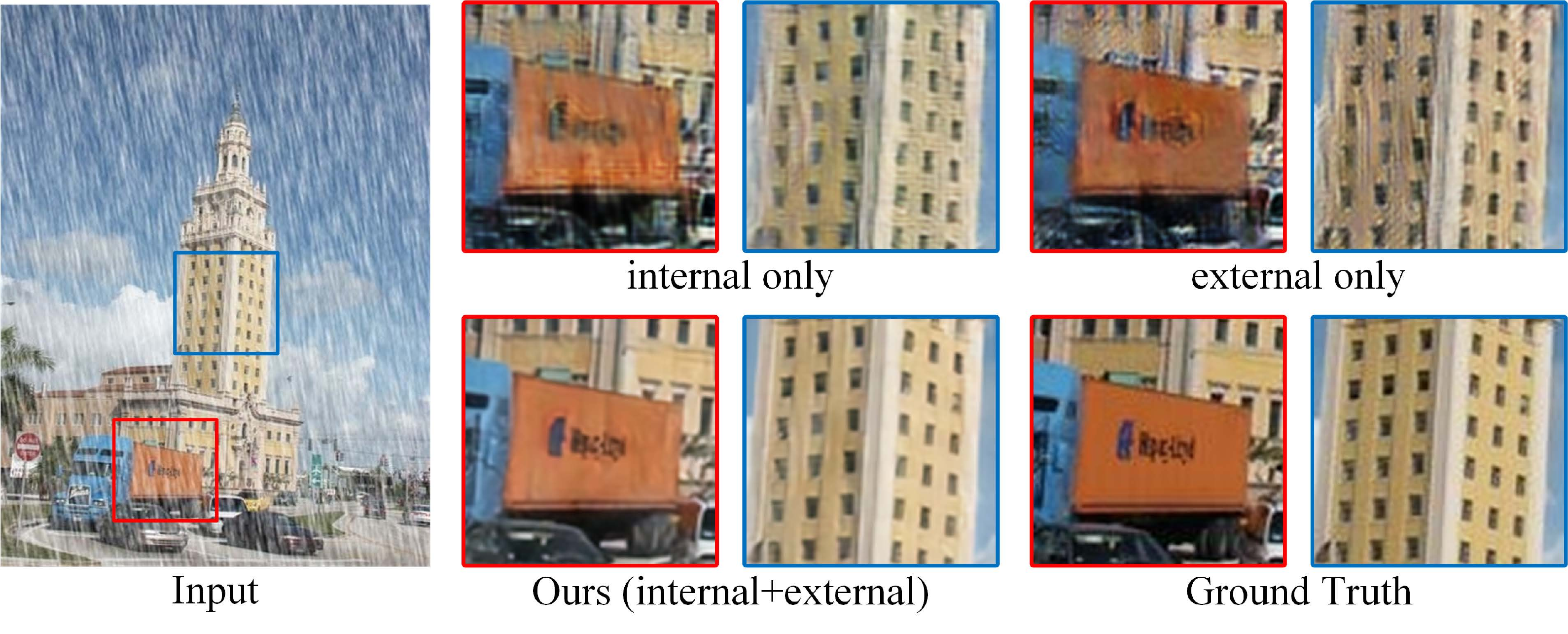}
	\vspace{-6mm}
	\caption{Qualitative analysis of negatives selection strategies.} 
	\label{fig7}
\end{figure}
\begin{figure}[!t]
	\centering
	\includegraphics[width=1.0\columnwidth]{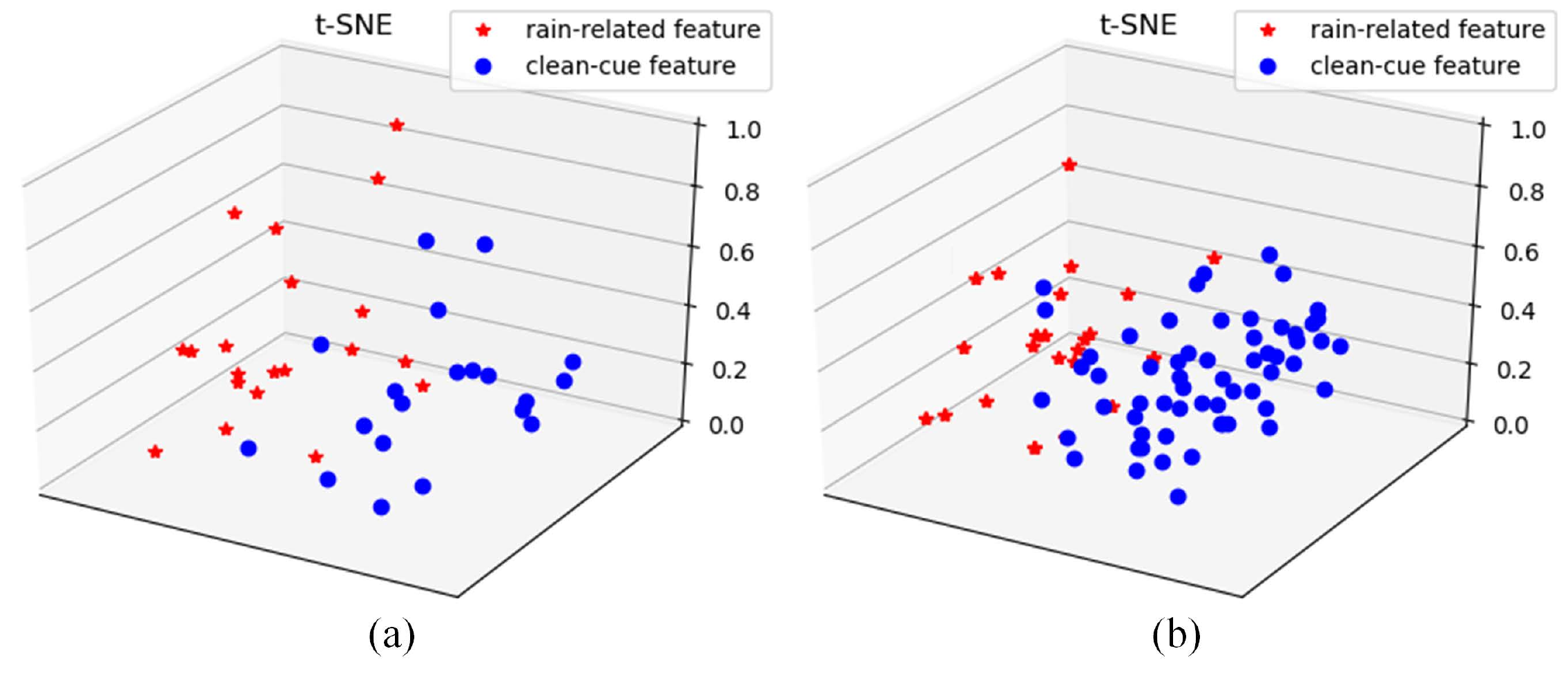}
	\vspace{-6mm}
	\caption{Visualization results of t-SNE. (a) w/o backward cycle; (b) w forward+backward cycle.} 
	\label{fig8}
\end{figure}

{\flushleft\textbf{Influences of Selecting Negatives}.} The most important design in contrastive learning is how to select the negatives. At each iteration, our developed DCD-GAN can generate four different stacks of features. Due to the advantages of dual learning setting, we concatenate the two stacks of features on both sides of the CGB together to provide more negative samples. Here, we explore the influence of this strategy by comparing using only 255 internal or 255 external negatives from one side. In Tab. \ref{table5}, the quantitative results on Test100 show that our strategy achieves the highest value of PSNR, while the method using only internal negatives obtains the best SSIM. However, zooming the color boxed in Fig. \ref{fig7}, we observe that the output quality of the comparison models are not as good as that of our strategy, that is, their derained results tend to make the image blur.

{\flushleft\textbf{Influences of Clean Exemplars}.} To further verify the influences of the exemplars from the unparied clean images, we remove the backward cycle as our comparison model. The quantitative results on Test100 are recorded in Tab. \ref{table6}. As one can see, the absence of the backward cycle substantially degrades results. In addition, we visualize the corresponding features using t-SNE \cite{van2008visualizing} in Fig. \ref{fig8}. Interestingly, we further observe that more clean-cue features are produced during the optimization. This is due to the backward cycle enriches clean exemplars, resulting in these unpaired data to better facilitate rain removal.

\begin{table}[t]
	\centering
	\caption{Ablation study for different loss functions. * indicates that $L_{c o l o r c y c}$ is replaced by $L_{c y c}$.}
	\vspace{-3mm}
	\resizebox{.95\columnwidth}{!}{
		\begin{tabular}{c|ccc}
			\hlinew{1.0pt}
			Model       & w/o $L_{a d v}$     & w/o $L_{c o l o r c y c}$     & w $L_{c y c}$*           \\
			PSNR / SSIM & 24.03 / 0.754 & 25.08 / 0.782 & 25.32 / 0.795 \\ \hline
			Model       & w/o $L_{c o n t}$         & w/o $L_{f r e q}$         & Ours          \\
			PSNR / SSIM & 22.84 / 0.731 & 25.40 / 0.799 & \textbf{25.61} / \textbf{0.813} \\ \hlinew{1.0pt}
	\end{tabular} }
	\vspace{-3mm}
	\label{table4}
\end{table}

\begin{table}[t]
	\centering
	\caption{Ablation study for different ways of selecting negatives.}
	\vspace{-3mm}
	\resizebox{.95\columnwidth}{!}{
		\begin{tabular}{c|c|c|c}
			\hlinew{1.0pt}
			Model       & internal only & external only & Ours (internal+external) \\ \hline
			PSNR / SSIM & 25.47 / \textbf{0.815} & 25.09 / 0.798 & \textbf{25.61} / 0.813   \\ \hlinew{1.0pt}
		\end{tabular}
	}	
	\label{table5}
	\vspace{-3mm}
\end{table}

\begin{table}[t]
	\centering
	\caption{Ablation study for different variants of our model.}
	\vspace{-3mm}
	\resizebox{.95\columnwidth}{!}{
		\begin{tabular}{c|c|c}
			\hlinew{1.0pt}
			Model     & w/o backward cycle & Ours (forward+backward cycle) \\ \hline
			PSNR / SSIM & 24.15 / 0.776        & \textbf{25.61} / \textbf{0.813}          \\ \hlinew{1.0pt}
		\end{tabular}
	}	
	\label{table6}
	\vspace{-3mm}
\end{table}

\begin{figure}[!t]
	\centering
	\includegraphics[width=1.0\columnwidth]{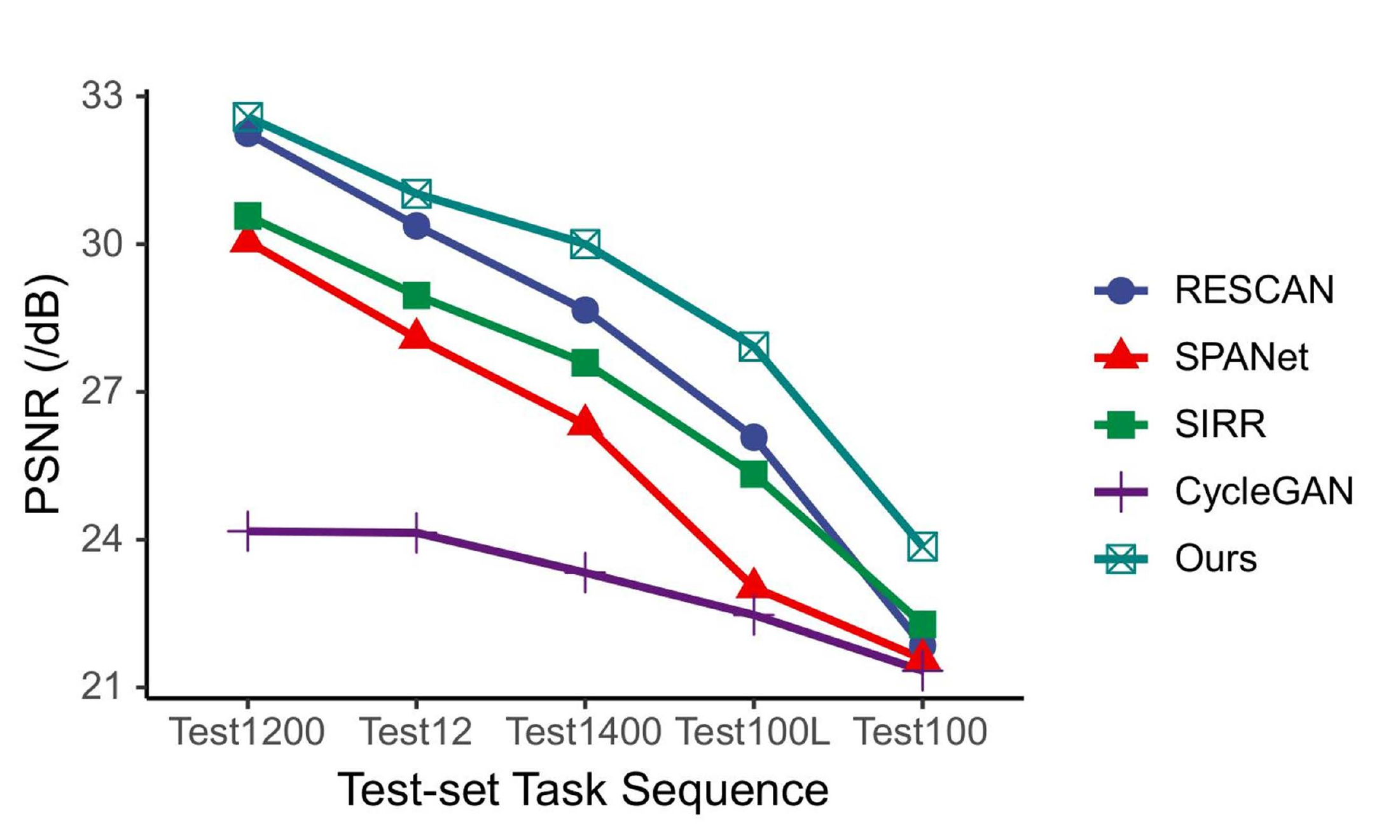}
	\vspace{-6mm}
	\caption{Quantitative analysis on cross-domain generalization.} 
	\label{fig9}
	\vspace{-3mm}
\end{figure}

\begin{figure*}[!t]
	\centering
	\includegraphics[width=0.95\textwidth]{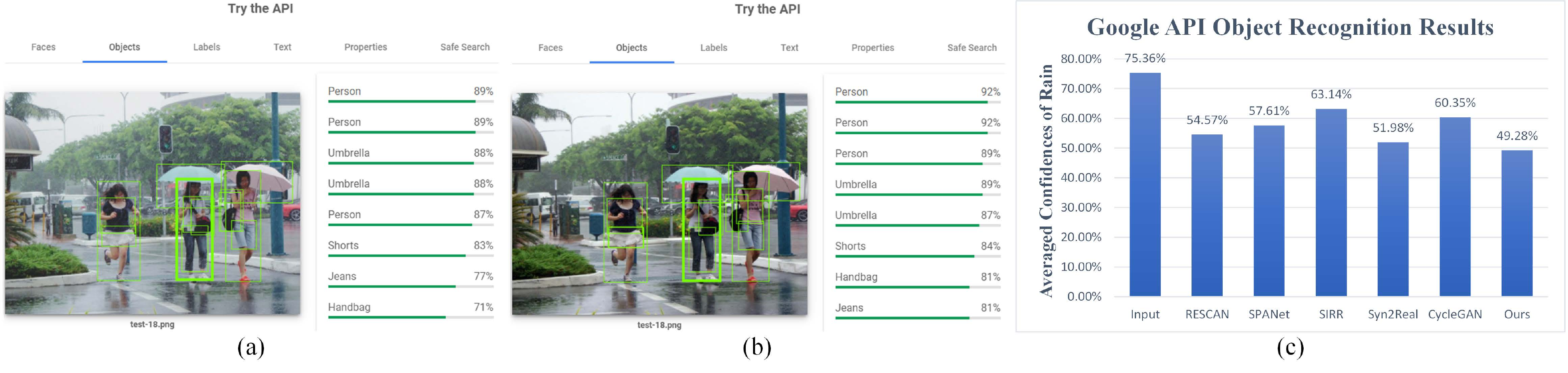} 
	\vspace{-4mm}
	\caption{Comparison results tested on the Google Vision API. (a-b) Object recognition results for the input rainy image and our derained image; (c) The averaged confidences in recognizing rain. Note that lower scores indicate better performance.}
	\vspace{-3mm}
	\label{fig10}
\end{figure*}

\begin{figure}[!t]
	\centering
	\includegraphics[width=0.95\columnwidth]{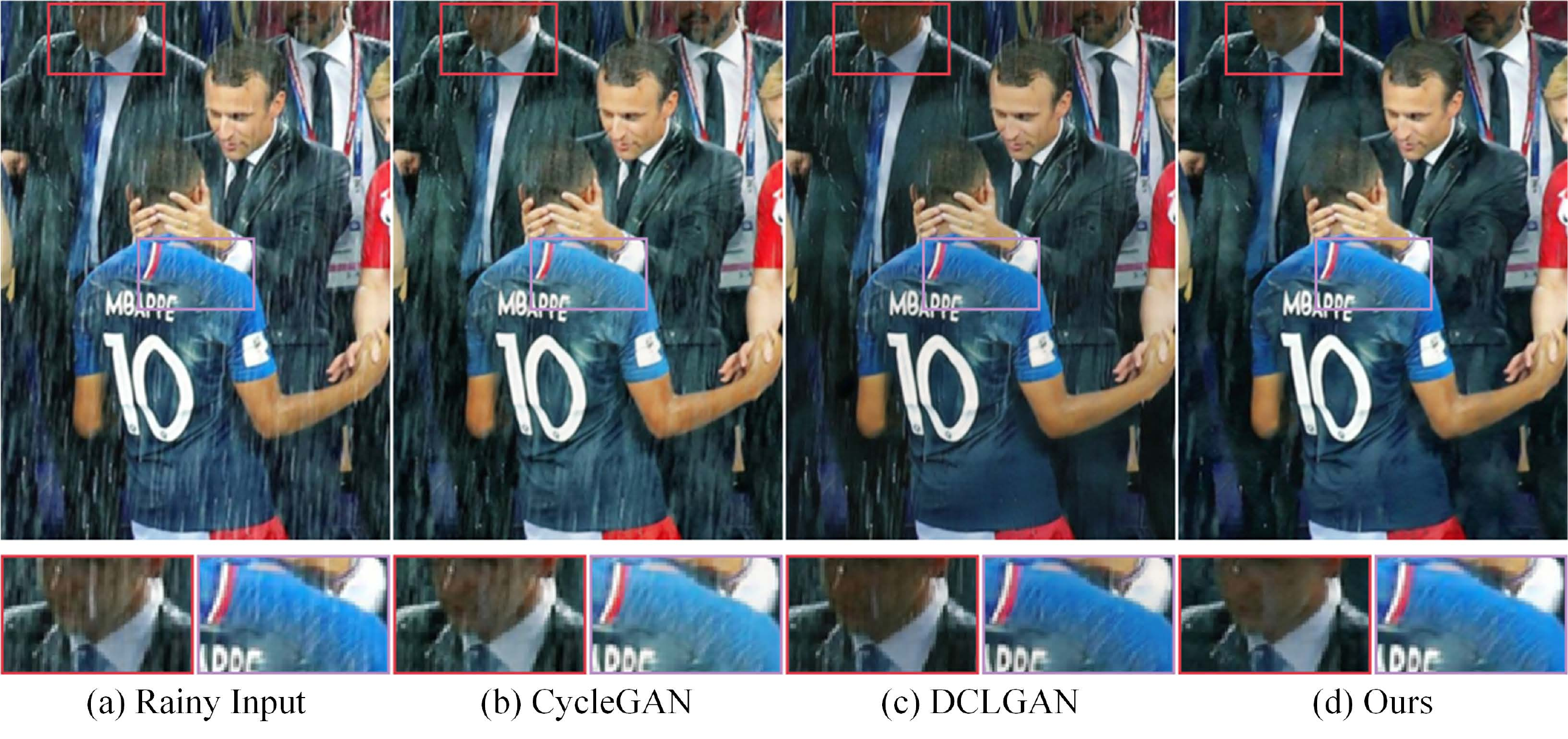}
	\vspace{-3mm}
	\caption{Comparison results when trained on the mixed dataset.} 
	\label{fig11}
	\vspace{-3mm}
\end{figure}

\subsection{Cross-Domain Generalizability Analysis}
We also conduct experiments to demonstrate that DCD-GAN can promote its cross-domain generalization. Specifically, the model trained on the DID-Data \cite{zhang2018density} is used to sequentially test the task sequence Test1200 \cite{zhang2018density} $\rightarrow$ Test12 \cite{li2016rain} $\rightarrow$ Test1400 \cite{fu2017removing} $\rightarrow$ Test100L \cite{yang2017deep} $\rightarrow$ Test100 \cite{zhang2019image}, and the result is reported in Fig. \ref{fig9}. It is evident that the performance of the fully-supervised approaches decreases dramatically due to domain variation. Thanks to the learned latent restoration in the deep feature space, our method could be easily adapted without requiring any paired data in the new domain, which greatly boosts its cross-domain deraining generalization.

\subsection{Discussions with the Closely-Related Methods}
To better solve the SID problem, we implement a mixed training manner. Here, we assemble a mixture of Rain800 \cite{zhang2019image} and 356 real rainy images (selected elaborately and cropped manually to highlight the rainy regions), which can provide more rain streak information for model learning. We note that \cite{zhu2017unpaired} introduces a cycle-consistent constraint for unpaired image-to-image translation. In this work, to better achieve SID, we further develop multiple CGBs to provide complementary constraints. Fig. \ref{fig11} (b) shows that CycleGAN fails to remove rain streaks, and it tends to blur the contents and cause color distortion. In contrast, we construct an effective contrastive regularization to explore the correlation of the latent space between the rainy and rain-free images, which thus leads to better derained results.

In addition, we also note that \cite{han2021single} recently develops a dual contrastive learning method for unsupervised image-to-image translation based on the CycleGAN framework. Different from this method learns domain-invariant information in complex image space, we focus on domain-specific properties in deep feature space. As displayed in Fig. \ref{fig11} (c) and (d), the recovery result of DCLGAN has some residual rain streaks, while our proposed DCD-GAN can deal with majority of rain streaks. Since the mapping in the low-dimensional feature space is in principle much easier to learn than in the high-dimensional image space \cite{wan2020bringing}, another benefit can be found is being good at avoiding unnecessary image structure distortion. Due to the introduction of real rainy images in the mixing training, it also supplies more negative samples to CGB to some extent. One more thing, cycle-consistent constraint and contrastive constraint share some commonalities \cite{ho2020contrastive}, which will be a double benefit for encouraging deraining performance.

\subsection{Application}
To examine whether our method benefits outdoor vision-based applications, e.g., object recognition, we apply the detection method, i.e., Google Vision API, to evaluate the derained results. Fig. \ref{fig10} (a) and (b) show that the recognition accuracy is improvement by using our derained output, suggesting that DCD-GAN can effectively improve the subsequent detection performance. Similar to \cite{deng2020detail,yi2021structure}, we test 30 sets of real-world rainy images and derained images of six different methods, as shown in Fig. \ref{fig10} (c). The confidence of rain indicates that the probability of rainy weather. As one can see, the averaged confidences in recognizing rain from our output results are significantly reduced, which indicates that DCD-GAN can restore high-quality images against other approaches. Therefore, it is meaningful to our developed unpaired deraining method to promote the practicality and scalability of real-world applications.

\section{Concluding Remarks}
In this paper, we have presented an effective unpaired learning, i.e., DCD-GAN, for SID. 
We incorporate the bidirectional translation branch and the contrastive guidance branch into a unified framework to achieve the joint training. Contrastive learning is first introduced into SID task to explore mutual features while distinguish the dissimilar ones in the deep feature space for better image restoration. 
Our approach enables the features from the unpaired clean exemplars to facilitate rain removal. Experimental results on synthetic and real rainy datasets considerably show that the effectiveness and generalization of our model. 
In future work, we will explore the potential of our proposed learning scheme in other unsupervised low-level vision tasks.

{\flushleft\emph{Limitation}.} 
Our method has limitation in SID performance when training on the small-scale datasets, due to contrastive learning tends to require a large number of sample pairs to achieve excellent performance.

\newpage

{\small
\bibliographystyle{ieee_fullname}
\bibliography{egbib}
}

\end{document}